# Region-wise stacking ensembles for estimating brain-age using MRI

Georgios Antonopoulos[1,2], Shammi More[1,2,3], Simon B. Eickhoff[1,2], Federico Raimondo[1,2], Kaustubh R. Patil[1,2]

[1] Institute of Neuroscience and Medicine (INM-7: Brain and Behaviour), Research Centre Jülich, Jülich, Germany
[2] Institute of Systems Neuroscience, Heinrich Heine University Düsseldorf, Düsseldorf, Germany
[3] Department of Bioinformatics, Fraunhofer Institute for Algorithms and Scientific Computing (SCAI), Sankt Augustin, Germany

## Highlights

- Brain-region models effectively fuse voxel-wise information
- Two-level stacking ensembles improve age prediction performance
- Regional age predictions provide novel insights into brain aging
- High performance without compromising privacy facilitates real-world applicability

## Abstract

Predictive modeling using structural magnetic resonance imaging (MRI) data is a prominent approach to study brain-aging. Machine learning algorithms and feature extraction methods have been employed to improve predictions and explore healthy and accelerated aging e.g. neurodegenerative and psychiatric disorders. The high-dimensional MRI data pose challenges to building generalizable and interpretable models as well as for data privacy. Common practices are resampling or averaging voxels within predefined parcels, which reduces anatomical specificity and biological interpretability as voxels within a region may differently relate to aging. Effectively, naive fusion by averaging can result in information loss and reduced accuracy.

We present a conceptually novel two-level stacking ensemble (SE) approach. The first level comprises regional models for predicting individuals' age based on voxel-wise information, fused by a second-level model yielding final predictions. Eight data fusion scenarios were explored using as input Gray matter volume (GMV) estimates from four datasets covering the adult lifespan. Performance, measured using mean absolute error (MAE), R2, correlation and prediction bias, showed that SE outperformed the region-wise averages. The best performance was obtained when first-level regional predictions were obtained as out-of-sample predictions on the application site with second-level models trained on independent and site-specific data (MAE=4.75 vs baseline regional mean GMV MAE=5.68). Performance improved as more datasets were used for training. First-level predictions showed improved and more robust aging signal providing new biological insights and enhanced data privacy. Overall, the SE improves accuracy compared to the baseline while preserving or enhancing data privacy.

## Introduction

The process of aging in humans is complex and it inevitably influences the brain, with negative consequences such as neurodegeneration which can lead to dementia and also is a mortality risk [1], [2]. The structural changes in the brain can be measured non-invasively via Magnetic Resonance Imaging (MRI) scans which capture the brain in volumetric form, comprising smaller volume elements called voxels. Each scan consists of hundreds of thousands of voxels. Following appropriate processing, it is possible to quantify the amount of specific brain tissues such as gray matter volume (GMV) at each voxel. This in turn permits in-depth study of distinct brain structures in relation to various cognitive, pathological and other physiological processes such as aging. Structural differences in the GMV have been reported between younger and older healthy individuals [3], [4], [5] along with continuous, widespread reduction in GMV observed from middle age onwards [6], [7], [8]. Additionally, an accelerated loss of both global and local GMV has been documented in neurodegenerative disorders [7], [9], [10], [11], [12], [13] such as the Alzheimer's disease (AD) and mild cognitive impairment (MCI) compared to normal aging.

Leveraging machine learning models to predict chronological age in healthy individuals using GMV aims to estimate the trajectory of healthy brain aging and through it unravel the links to neurodegenerative, psychiatric, and other disorders. Such MRI-derived age-prediction has been shown to be a reliable proxy for overall health [10], [14], [11]. An elevated brain age stands as an important risk factor for various neurodegenerative and psychiatric disorders [12], [13], [15], [16], [17]. Thus, age prediction models can facilitate early detection of health risk and facilitate effective prevention and treatment strategies. The success of this approach hinges on accurate and biologically meaningful models. Furthermore, sharing MRI data or even its derivatives can lead to privacy issues [18]. Hence it is important that new brain age estimation methods respect individuals' privacy. Furthermore, generalizing across data from multiple scanners remains challenging due to systematic biases. To this end, various machine learning algorithms and GMV-derived features have been tested aiming at providing more accurate predictions as well as insights into healthy brain aging. For instance, Relevance Vector Regression (RVR) outperformed Support Vector Regression (SVR) when trained on downsampled voxels -with and without feature selection, yielding MAE=5 years in healthy subjects and MAE=10 years in AD [19]. Gaussian Process Regression using structural MRIs was efficient in estimating mortality via brain age [1] and outperformed RVR, Kernel Ridge Regression and LASSO [6]. Additionally, various deep neural network models have been proposed such as convolutional neural networks [13], [19], [20], DenseNet [21], lightweight deep learning model, Simple Fully Convolutional Network (SFCN) [22] and other variations [23], [25].

In terms of GMV features, previous studies have used voxel-wise GMV [1], [2], [26], [27], feature reduction methods such as principal component analysis and non-negative matrix factorization [25], [26] as well as regional mean values [30], [31]. The voxel-based approach is encumbered by the "curse of dimensionality", high computational demands due to the high number of voxels and high intersubject variability. On the other hand, the use of regional mean GMV is more conducive for machine learning. This simple and efficient type of information fusion can enhance robustness and statistical power mitigating the impact of noise or variability within individual voxels. Additionally, use of a meaningful parcellation scheme increases sensitivity for detecting regional changes, yielding results that are easier to interpret from a neuroanatomical perspective and can provide critical priors for studying brain basis of behavior and disease [32]. Effectively, calculating regional mean of GMV as a form of information fusion might be too naive as it weighs all the voxels equally regardless of their reliability and relevance to the problem at hand.

Further challenges to real-world use of brain age models are posed by data heterogeneity and data privacy issues. It is desirable that the models should work effectively on test data collected from a new scanner from which no or only limited data is available for training. However, scanner-induced systematic differences make this challenging [33], [34]. It is also desirable that data is utilized and shared in a way that preserves privacy, a crucial consideration in healthcare and medicine [35]. Any use of personal and medical data should ensure it's ethical for research and treatment, and comply with legal obligations. For instance, raw MRI data is enough for face identification of the subjects [36] as well as information such as MRI-derived connectomes are unique to individuals like a fingerprint promoting identification [37]. Subjects can be identified based on their regional GMV, even across different preprocessing pipelines, indicating that sharing preprocessed data still poses the risk of privacy violation [34].

To better utilize the information of each voxel, while managing the very high dimensionality of the data, we propose the use of stacking generalization or stacking ensemble (SE) models [38] which effectively mitigate overall bias and variance, and have shown promise in multiple domains [39], [40], [41]. The SE framework is based on weighted voting of predictions derived from various models (referred to as generalizers in [38]). Specifically, we propose a variation of SE in which first level models are trained on the voxel-wise data from each brain region. In contrast to the uniform weighting, as done in conventional models that use the mean for each region, the first level models weight the contribution of each voxel according to how it captures the aging signal. This approach can effectively handle regions in which the voxels have differing signal to noise ratio. A second level model is then trained using the predictions of the first level models to obtain the final prediction. In addition, the predictions of the first level models are more aligned with the target, in our case age. This allows for better fusion of data across datasets/scanners, in effect mitigating the scanner differences and allowing to combine datasets to train more accurate and generalizable models. In other words, pooling age predictions of first level models is likely to incur lower bias than pooling regional mean GMV. Overall, we expect that this more nuanced and informative fusion of voxel and cross-dataset information contrary to the conventional mean approach will potentially lead to more accurate and biologically meaningful models.
The proposed SE framework, to some degree, addresses also the privacy issues. By using the output of the first level, i.e. regional age predictions, for cross-site predictions and sharing our framework promotes interoperability and privacy.

While SE models have been previously used in age-prediction for combining the predictions of various modalities [11], [42], they have not been used for region-wise structural MRI models. Popescu et al [27] utilized structural MRI features to predict age for each brain region, however, they did not combine regional predictions in a meta model. In this study we use a SE framework to enhance brain-age predictions and provide an alternative and more robust biological representation of the contribution of brain regions in healthy aging. We employed structural MRI scans of healthy individuals from four large open datasets covering the adult lifespan. We systematically explored various ways to fuse data from different sites during training and testing. Specifically, we performed cross-site predictions, by keeping a test site separate from the training sets, while ensuring consistent training sample sizes across all set ups. Furthermore, we also explored different methods for conducting out-of-sample (OOS) predictions with and without site-specific models. We compared the conventional baseline utilizing the mean GMV for each region against using the SE framework in all those set ups. Additionally, we assessed the impact of the number of datasets available for training and examined their stability on data coming from different sites. At the current stage we did not perform any comparison between SE models and the state-of-the art (SOTA) methods. Our goal is to demonstrate the theoretical and practical promise of SE models in this specific context rather than to immediately outperform highly tuned, task-specific models. Our systematic analysis provides valuable insights regarding the efficacy of our proposed stacking ensemble framework in the domain of brain age prediction, paving the path for clinical applications.

## Material and methods

### Datasets and preprocessing

We used T1-weighted MRI scans of healthy individuals (total N=2926, covering the whole adult lifespan (18-88 years), coming from 4 open datasets: the Cambridge Center for Ageing and Neuroscience (CamCAN, *N*=650, mean age=54±18.6, min-max=18-88) [43], Information eXtraction from Images (IXI, *N*=562, mean age=48.7±16.45, min-max=20-86) (https://brain-development.org/ixi-dataset/), the enhanced Nathan Kline Institute-Rockland Sample (eNKI, *N*=597, mean age=48.2±18.5, min-max=18-85) [44] and the 1000 brains study (1000Brains, N=1117, mean age=61.8±12.4, min-max=21-85) [45].

We processed the T1 scans to extract modulated GMV in the MNI space for each subject, using Computational Anatomy Toolbox (CAT) version 12.8 [23]. To ensure accurate normalization and segmentation, initial affine registration of T1w images was done with higher than default accuracy (accstr=0.8). After bias field correction and tissue class segmentation, optimized Geodesic Shooting [46] was used for normalization (regstr=1). We used 1.5 mm Geodesic Shooting templates and outputted 1.5 mm isotropic images. The normalized GM segments were then modulated for linear and non-linear transformations yielding estimated GMV for each voxel (N=399184 voxels in total).

### The SE model training

In this study we implemented SE with two levels, denoted as L0 and L1. For both levels, we utilized GLMnet (elastic net) [47] due its ability to handle multicollinearity in data, such as the voxels of structural MRI, efficiently. Elastic net regression combines ridge regression (L2 regularization) and LASSO (L1 regularization), which facilitates dealing with multicollinearity by penalizing large coefficients and promoting simpler models. We used the glmnet package in R (version R-4.1.0) that incorporates an internal process for hyperparameter tuning. By default it optimizes the regularization parameter lambda (*λ*) along with the mixing parameter alpha (α) using 'adaptive resampling'. Features with near-zero variance were identified and removed. The remaining features underwent centering -removing the mean-, and scaling -dividing with the standard deviation.

At the first level L0, GM voxels were grouped into 873 regions encompassing cortical, subcortical and cerebellar regions, using a parcellation atlas [31] with 800 cortical regions from the Schaefer atlas [48], 36 subcortical regions from the Brainnetome Atlas [49] and 37 cerebellar regions [50]. We chose this specific granularity to retain anatomical specificity. Nevertheless, other options could be also explored. We employed a 3-fold cross-validation scheme on the training data to generate voxel-based and out-of-sample age predictions for each region. These predictions (873 per subject) were then used as inputs to train the second level L1 model. For application on new unseen samples, a final L0 model (GLMnet) per region was trained on the complete training set. The L1 GLMnet model was trained on the L0 models' OOS predictions from the CV and provides the final age prediction.

## Data split and training setups

In our experiments, we estimated models' performance using Leave-One-Site-Out (LOSO) validation. The LOSO set up mimics the scenario when the models are applied to data from a new scanner not available during training, thus it provides a more accurate estimation of model's ability to generalize across different scanners or sites. When using multiple datasets for training, the training data or models can be combined in various ways. Data from different sites can be either pooled and used as if they are coming from one source for training L0 and L1 models. Alternatively, data from each site can be used separately to train both in L0 and L1 models creating site-specific SE models and the predictions of L1 models can be then averaged to obtain the final prediction. In-between set ups are also possible, for instance L0 predictions can be pooled or averaged prior to training a L1 model. We implemented 8 such set-ups as presented in Table 1. We also examined the impact of training with one or multiple different sites, always testing in an unseen site.

Additionally, beyond the aforementioned set-ups, we tested the scenario where a center or a clinic cannot share raw data from patients. Thus, L0 regional predictions could be estimated directly in the test site (from the test set), using a k-fold CV scheme to get out-of-sample (OOS) predictions and pass these predictions to L1 models, trained either in pooled or site-specific data. Note that the L1 models did not include data from the test site for training. We tested this scenario, using a 3-fold CV to get L0 predictions from the test set.
To benchmark against the conventional standard, we computed the mean of GMV for each region and trained a GLMnet model to predict age. This is equivalent to replacing each of the 873 first level models with an averaging function. We implemented both pooling as well as different set ups of data/model fusion for the GMV mean data. In order to have a fair comparison with the set-up where we use the test set in a 3-fold scheme in L0, we also implemented an approach for the mean GMV models, where we trained models by pooling the training data sets with the two training folds from the test site.

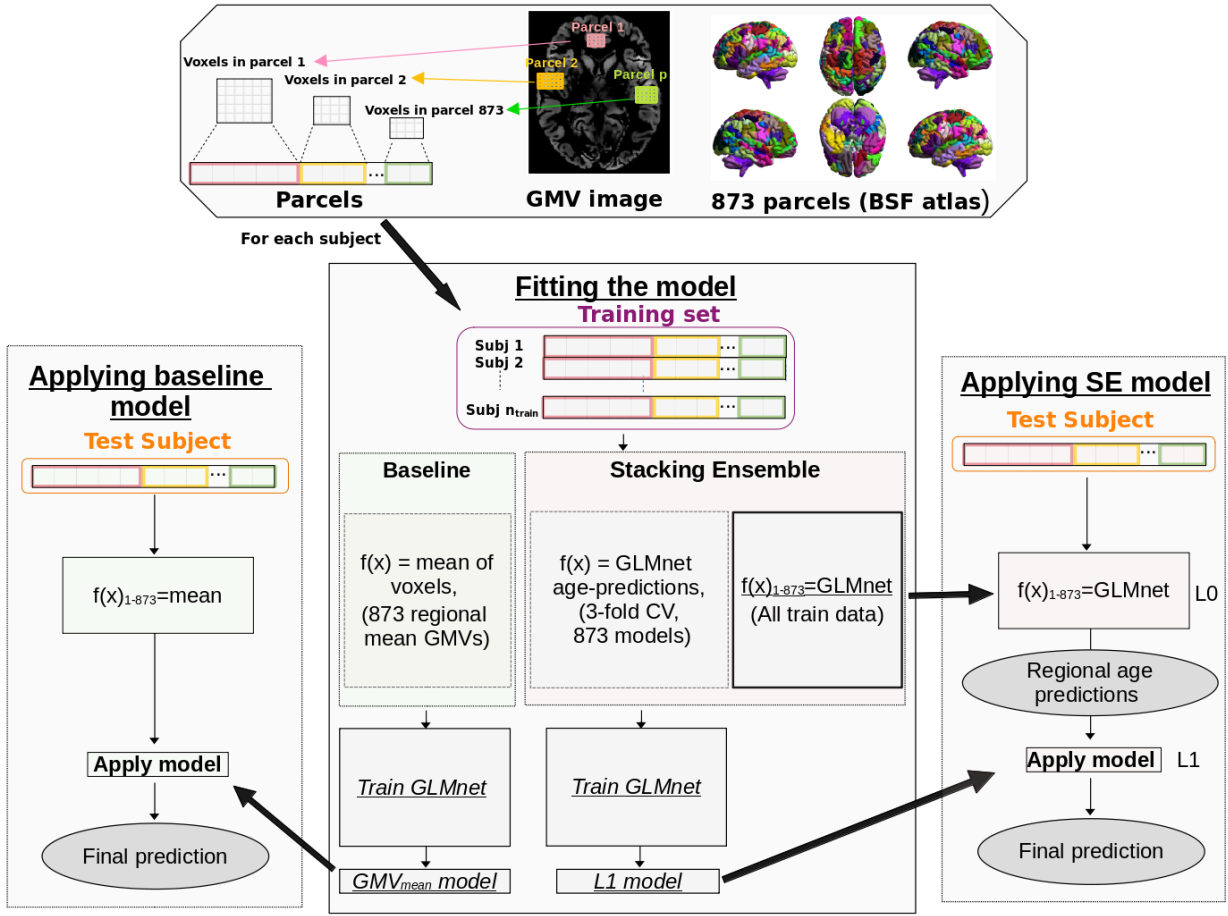

Figure 1: **Illustration of standard age-prediction process**. Left: Current standards perform a regional mean of GMV in level-0 (f(Vi)=mean( voxels in region i)) and with those they train a model in Level-1. In SE models voxels in each region are used to train a model to predict age, in a K-fold scheme, and the predictions are used to train the Level-1 model, which is used to estimate the final prediction. Right: In L0 the trained models (or local averaging) are applied and their output is provided as inputs for the L1 model, which makes the final prediction.

| Train datasets | L0 operation | L0 train data/output | L1 operation | L1 train data/output | Notation |
|---|---|---|---|---|---|
| per site | $f_{1-873}(x)$=mean() | per site | $f_{L1}(x)$=GLMnet per site, mean sites | three outputs Mean() | **GMV$_S$L1$_S$** |
| | $f_{1-873}(x)$=GLMnet | per site | $f_{L1}(x)$=GLMnet per site, mean sites | three outputs Mean() | **PredL0$_S$L1$_S$** |
| | OOS on test set | per site | $f_{L1}(x)$=GLMnet per site, mean sites | three outputs Mean() | **OOSPred$_S$L1$_S$** |
| | $f_{1-873}(x)$=mean() | pool preds | $f_{L1}(x)$=GLMnet on pooled L0 preds | one output | **GMV$_P$L1$_P$** * |
| | $f_{1-873}(x)$=GLMnet | per site | $f_{L1}(x)$=GLMnet on pooled L0 | one output | **PredL0$_S$L1$_P$** |
| | OOS on test set | per site | $f_{L1}(x)$=GLMnet on pooled L0 | one output | **OOSPred$_S$L1$_P$** |
| pooled | $f_{1-873}(x)$=mean() | pool preds | $f_{L1}(x)$=GLMnet | one output | **GMV$_P$L1$_P$** * |
| | $f_{1-873}(x)$=GLMnet | pool preds | $f_{L1}(x)$=GLMnet | one output | **PredL0$_P$L1$_P$** |
| | OOS on test set | pool preds | $f_{L1}(x)$=GLMnet | one output | **OOSPred$_P$L1$_P$** |

Table 1. All the setups that we compared, given the possible combinations that can occur when creating site-specific models or pooling data from all sites in the two levels of SE models.
* These models are the exact same, as the L0 is local averaging and has nothing to do with the train data.

The utility of using additional datasets was evaluated by comparing the performances of the different setups when using a different number of training data sets. For this, we examined the cases where one, two and three datasets were used to train the models in all possible combinations occurring from our four datasets.

For performance evaluation, we calculated the mean absolute error (MAE), coefficient of determination ($R^2$), Pearson's correlation between real age and predicted age (r), and prediction bias. The prediction bias is calculated as the correlation between the actual ages and the prediction errors (real age - predicted age) and it expresses the tendency of a model to provide predictions closer to the mean, meaning that younger subjects are predicted as older and older are predicted as younger [33], [51]. Although it is not a standard performance metric, it is important for brain-age prediction problems. These metrics are widely used and well described in MRI-based age-prediction problems [52].

## Biological insights

To understand biological aging it is important to identify regions that are strongly and robustly correlated with the real age. Although correlation doesn't imply causation, such insights can shed light on the brain-aging process. To this end, we calculated Pearson's correlation of real age with L0 predictions and mean regional GMV across subjects. These values were compared to identify the more robust approach providing biological insights. Projecting those correlations back to the brain space reveals regions strongly and robustly associated with aging. We ensured the stability of those results by evaluating all four datasets to show that the insights gained are not dataset dependent. This comprehensive analysis was aimed to provide a holistic perspective on our method's robustness and stability.

## Data privacy

To assess how SE compares to conventional methods in data anonymity and privacy, we conducted a multiclass classification task to predict the dataset of origin. Identifying the datasets is a step closer to identifying the subject itself and thus this task serves as a proxy for gauging the potential for privacy violation. For this task, we employed L0 regional age predictions as features and juxtaposed the dataset prediction performance with that achieved using regional mean gray matter volume (GMV). We performed model selection with hyperparameter tuning for both sets of features. The tested models included logistic regression, linear Support Vector Machine (SVM), SVM with radial basis function kernel, Random Forest (RF). The exact hyperparameter space is presented in the Supplementary Table 1. The accuracy of the optimal model serves as a metric, where higher accuracy indicates better preservation of dataset-specific information in the corresponding feature space. Conversely, lower accuracy in the dataset prediction task suggests a higher level of privacy. This analysis permits to directly compare how SE preserves data anonymity in contrast to regional GMV.

# Results

## Performance

The performance of different set ups was evaluated using leave-one-site-out (LOSO) scheme. We calculated MAE, $R^2$, Pearson's correlation between predicted and real values and prediction bias. Overall, SE frameworks showed better performance. Specifically, the highest performances were observed for the setups where L0 predictions were obtained from the test site: **OOSPred$_s$L1$_p$** (average MAE=4.75), closely followed by **OOSPred$_s$L1$_s$** (MAE=4.9). Setups using pooled L0 predictions to train a single L1 model, independently of how L0 was trained, i.e. **PredL0$_p$L1$_p$** and **PredL0$_s$L1$_p$**, showed MAE=5.1. When using mean region-wise GMV, the model trained using the training data sets together with the 2 training folds from the test site, **GMV$_p$L1$_p$ext**, showed MAE=5.7. Performance was worse for the other two region-wise mean GMV models trained in three datasets, with the L1p setup (pooled predictions from L0 to train L1) being slightly better compared to L1s (train L1 models from different sites), MAE=6.2 and MAE=6.7 respectively. The performance ranking was the same for $R^2$, slightly changed in Pearson's r, where the best model was the one where both L0 and L1 models were trained on pooled data. The performance of the models regarding age bias was rather different. The lowest bias was found for **GMV$_p$L1$_p$** followed by **PredL0$_p$L1$_P$** and **OOSPred$_s$L1$_P$** (b= -0.41, b=-0.43 and b=-0.44, respectively). Detailed results can be found in Supplementary material Table 3.

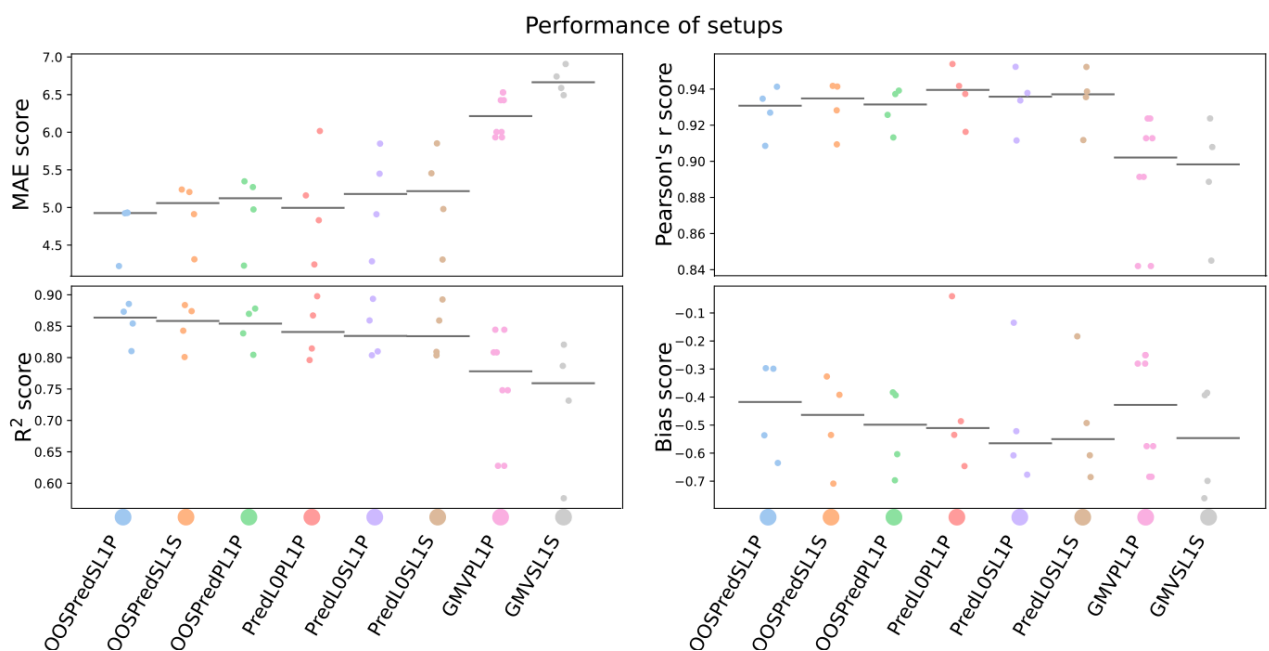

Figure 2: **Performance of SE models and GMV approaches for chronological age prediction, across four metrics**. Performance in terms of MAE (upper left), Pearson's correlation (upper right), $R^2$ (bottom left) and bias (bottom right) for all setups. Results were extracted by using three datasets for training and one for testing in all possible combinations. Setups that perform OOS prediction on the test set yield better performance in all metrics except for bias.

All setups exhibited significant performance improvement with the inclusion of more datasets in the training process. Higher improvements in performance were observed when transitioning from one to two datasets, compared to the increase from two to three. This trend underscores the positive impact of creating the training data with multiple datasets, particularly in the initial stages of model development.

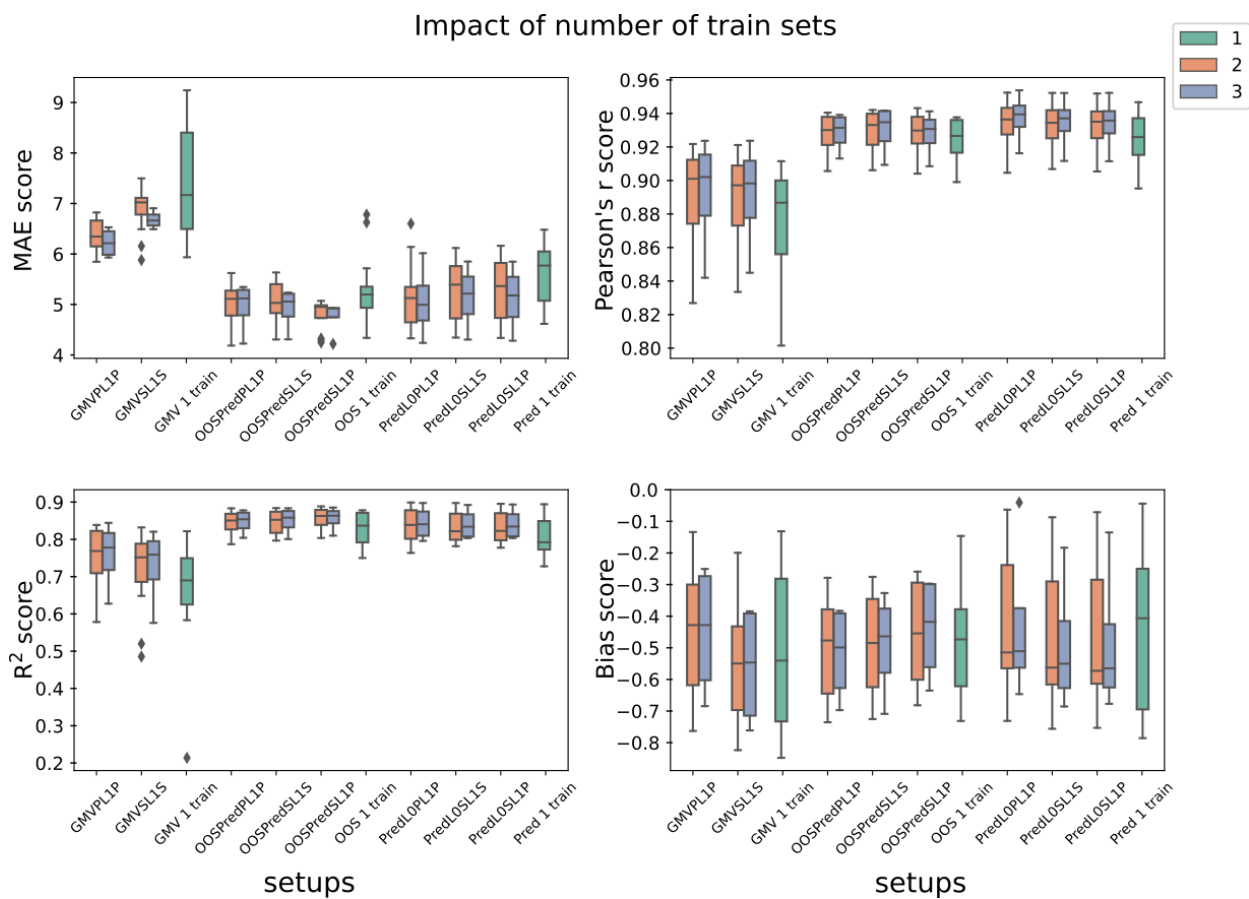

Figure 3: **Comparison of setups' performances across different numbers of train datasets**. The impact of the number of sites in the training process in terms of MAE (upper left), Pearson's correlation between predicted and real values (upper right), $R^2$ (bottom left) and bias (bottom right). Increase in the number of training sets yields better performance for MAE and $R^2$. Pearsons's correlation demonstrates small improvement and bias seems to be dependent on the setup.

## Biological insights

Correlation values between mean regional GMVs and chronological age across subjects is commonly used for identifying aging-related brain regions. Similar insights from SE models can be derived using the regional age-predictions provided by the L0 models. The results showed that regional predictions of the L0 models exhibit a stronger alignment to age (Figure 4). Correlation coefficients' mean (mean of absolute values) for L0 predictions-age was $r_{mean}=0.6$ and $r_{mean}=0.32$ respectively for GMV-age, yielding significant difference ($p<<0.01e^{13}$). Frontal and cerebellar areas were highlighted in both methods, notably stronger though in SE models. Remarkably, subcortical areas which exhibited positive mean GMV-age correlation, demonstrated a more pronounced and positive association in SE models. Specifically, four regions had positive GMV-age correlation for all datasets, with the mean across datasets ranging from r=0.12 to r=0.29. In comparison, the corresponding L0 predicted–age to real–age correlations were notably higher, ranging from r=0.62 to r=0.8.

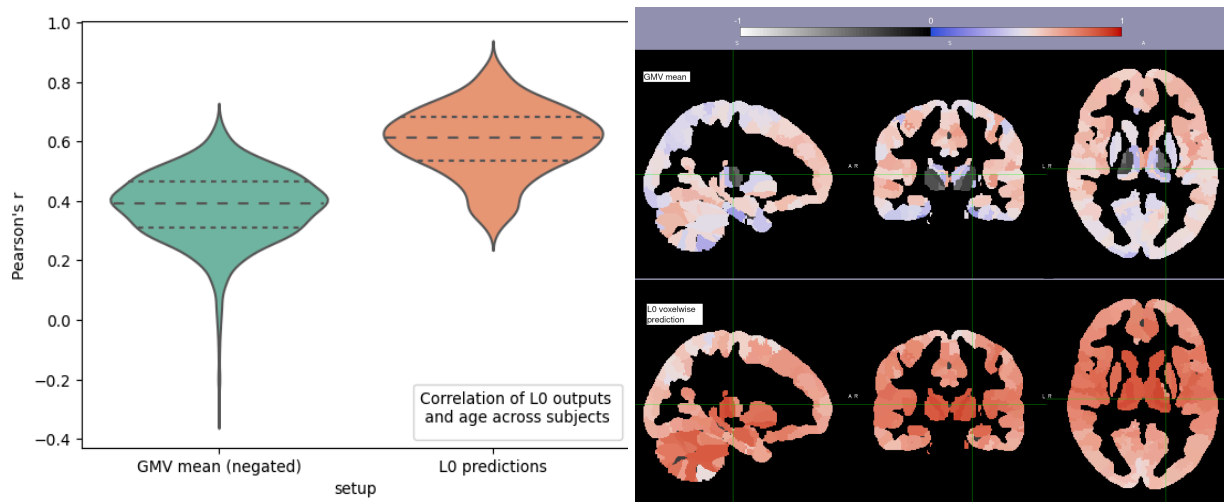

Figure 4: **Regional GMV-age and L0 predictions-age Pearson's correlations, presented in violin plots and projected in the brain**. The left panel illustrates Pearson's correlations, across subjects, between age and negated mean GMV per region (green), and that with L0 voxelwise predictions per region (orange). L0 predictions were more correlated to age, but more importantly all regions are positively correlated to age. The right panel demonstrates the same correlation values in the brain space. Positive correlation for GMV is found in the subcortical areas, where the correlation of age-L0 predictions appears to be very high.

Next, we sought to compare the stability of the biological insights provided by the two methods across different datasets. To this end, we calculated the correlation values between age and L0 regional age OOS-predictions across datasets and compared it to the correlation values calculated between age and regional mean GMV. Correlations between age and regional L0 age predictions demonstrated higher consistency across all datasets compared to the correlations of age and mean GMV (Figure 5). Cortical regions exhibited an average L0 predictions-age correlation of *r*=0.54, while the GMV-age correlation was negative at *r*=-0.4. In subcortical regions, the mean L0 predictions-age correlation was *r*=0.75, with the mean GMV-age correlation (absolute values) at *r*=0.35. Interestingly, some regions showed positive GMV-real age correlations; for instance, some thalamic regions had correlations of *r*=0.28 and *r*=0.29 when L0 predictions and real age correlations in these regions were *r*=0.8 and *r*=0.62, respectively. For the cerebellum, the mean L0 predictions-age correlation was *r*=0.60, while the GMV-age mean was *r*=-0.38.

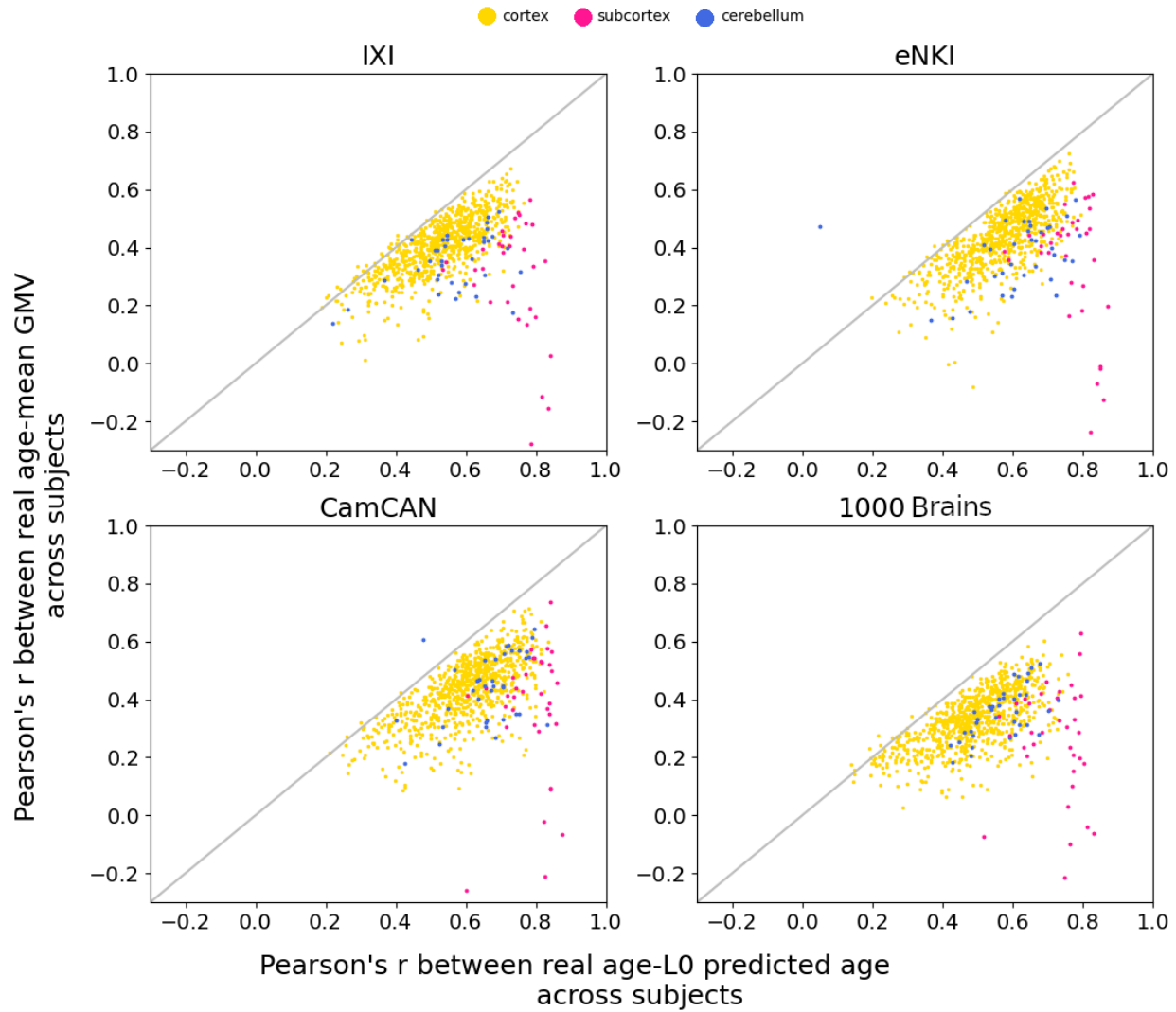

Figure 5: **Regional GMV-age and L0 predictions-age Pearson's correlations across datasets.** Pearson correlation across subjects was estimated between regional predicted age and real age, as well as between mean regional GMV and real age. The four panels show these correlations plotted against each other for each region for four datasets. Cortical regions are depicted in yellow, subcortical in pink and cerebellar in blue. We observed that SE L0 predictions to be more aligned with the real age indicating that it better captures the aging process. Here we demonstrate only the cases when one dataset was used for training.

### Stability across datasets

To mitigate the possibility that the associations between subjects' age and L0 predictions or GMV, are driven by datasets' idiosyncratic properties, we computed the correlation values separately for each dataset. The results (Figure 6) consistently exhibited the same pattern across datasets, affirming the overall stronger association of L0 (mean=0.86 ranging from 0.79 to 0.91) regional age predictions with age compared to mean regional GMV (mean=0.81 ranging from 0.75 to 0.9).

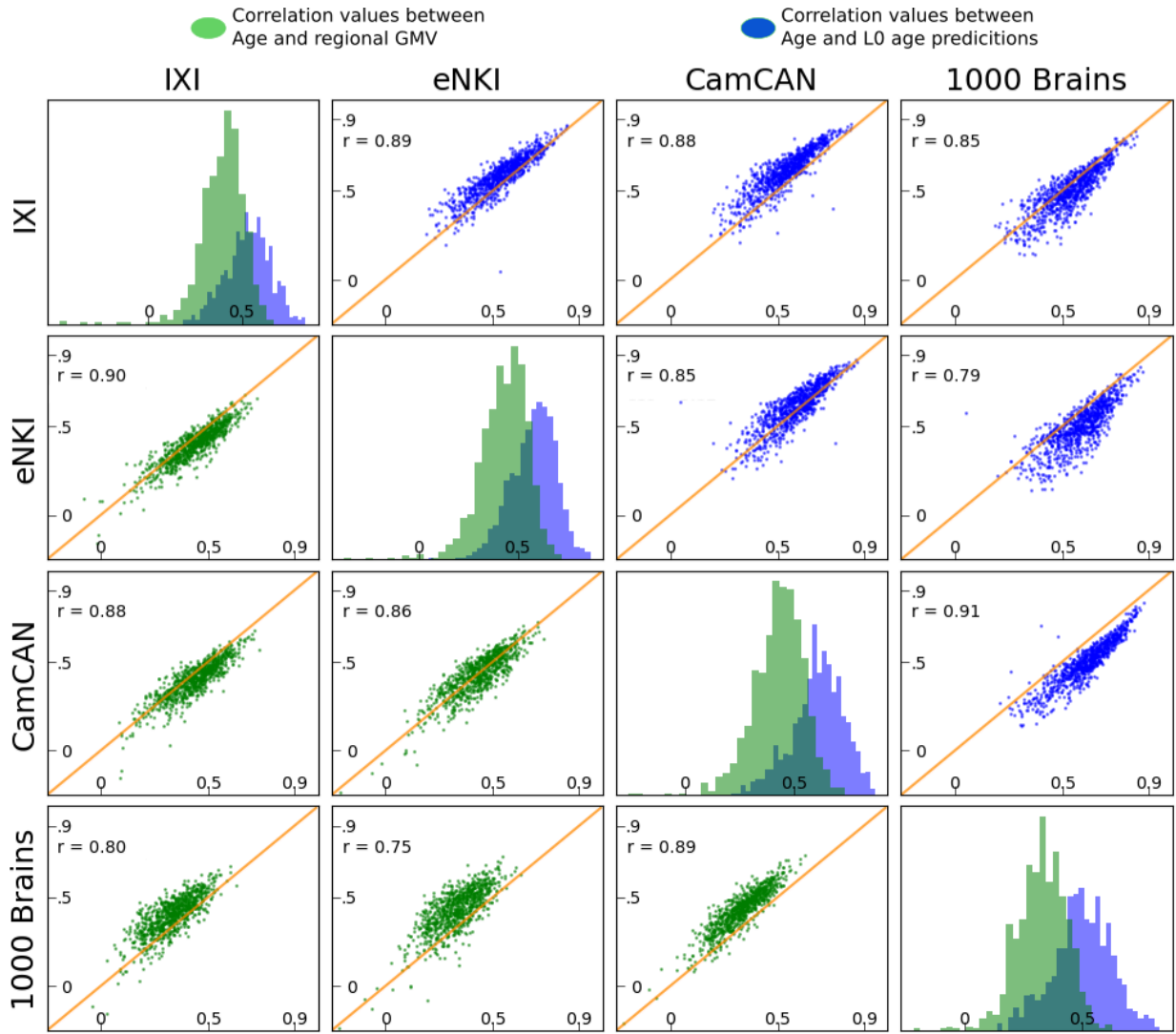

Figure 6: **Dataset based regional GMV-age and L0 predictions-age Pearson's correlations plotted against other datasets.** Pearson correlation was estimated between regional predicted age and real age, as well as between mean regional GMV and real age, across subjects. The scatter plots show these correlations against each other for each region. The plots show that SE L0 predictions are more aligned with the real age and can thus provide a better biological insight of the aging process in healthy individuals. Here we demonstrate the predictions obtained via a 3-fold CV scheme for each dataset.

## Privacy

For estimating whether SE can enhance the data privacy compared to the baseline of using mean GMV, we performed a multiclass classification task. The objective here was to predict the dataset of origin for each subject using either the regional L0 age predictions - out-of-sample estimations within each dataset-, or regional mean GMV as features. We performed a nested 5-fold cross validation with model selection and hyperparameter tuning happening within the inner CV. For both feature spaces linear SVM was chosen resulting in balanced accuracy $ACC_{bal}$= 0.63 SE L0 predictions and $ACC_{bal}$= 0.87 for mean GMV. The GMV model showed an overall higher confusion while the L0 predictions model was biased towards two sites (Figure 7).

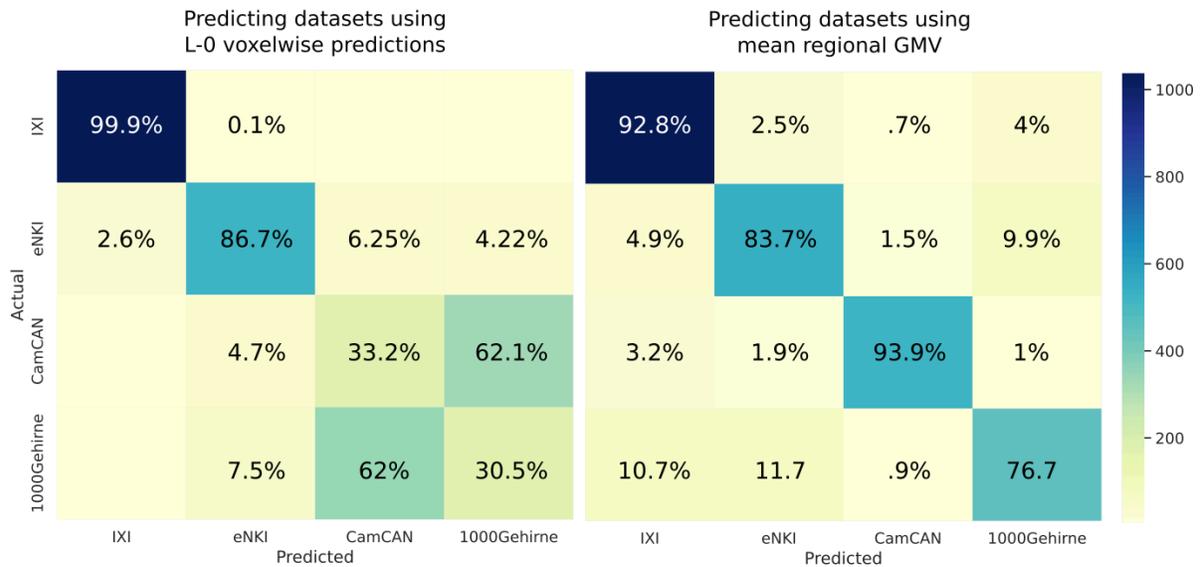

Figure 7: **Results of dataset prediction when GMV and L0 predictions are provided as features.** Confusion matrices for multilabel classification task of dataset of origin prediction with L0 regional predictions (left) and regional GMV (right). A high dataset prediction performance indicates a higher potential for identification of individuals.

## Discussion

Understanding and modelling healthy brain aging is crucial for developing individualized precision methods for various neurodegenerative and pathological brain disorders. Previous work has shown that advanced MRI-derived brain-age, i.e. a higher predicted age than actual age, is associated with neurodegenerative, psychiatric and other diseases [12], [13], [15], [16], [17]. Nevertheless, several important aspects need to be refined prior to integrating brain-age prediction into clinical practice: 1) improved performance and robustness, 2) interpretability and biological insights from prediction models, and 3) improved data privacy. In this context, we propose a two-level stacking model for MRI-based age-prediction. This novel approach performs voxel-based age predictions for each predefined brain region in its first level, which are then fused by a second-level model. Compared to the oftused baseline approach of averaging regional voxels' GMV, our approach offers a more sophisticated data fusion. Contrary to models using regional mean GMV, SE models make better use of the voxel-wise information. Our numerical experiments using large datasets with a wide age range suggest that the proposed SE model provides more reliable and accurate results.

Furthermore, datasets can be fused in different ways forming variations of SE models. Therefore, we tested setups differing in the way that training datasets were combined at both levels of the SE. Specifically, using a LOSO scheme, we evaluated three setups: pooling data from different sites, training per site and averaging results or performing OOS predictions in the first level using the test dataset. First we compared the performance of all SE setups and models to the standard approach of using regional mean GMV, in terms of MAE, $R^2$, Pearson's correlation and age bias. Overall, SE setups outperformed the regional mean GMV models in terms of MAE, $R^2$ and Pearson's correlation. Notably, all SE setups using a part of the test site data for L0 models to obtain out-of-sample predictions performed better than other setups in terms of MAE and $R^2$ (Suppl. Table 2), with those using regional GMV being the worst. In terms of Pearson's correlation, all SE setups showed r=0.93 (rounded in the second decimal digit) except for $PredL_0PL_1P$ (pooling data for both L0 and L1) that had a slightly better performance (r=0.94), and all outperformed GMV setups (r=0.89-0.9). This way of obtaining OOS L0 predictions effectively models the idiosyncrasies of the specific dataset. Additionally it also offers the advantage that centers/clinics do not need to share their raw data but only L0 predictions. In this scenario, the test sites can benefit from SE models (specifically L1) trained on publically available data while only using their own data for L0 models. In terms of the age bias, the results were more mixed. The best and worst bias were both setups using GMV, $GMVPL_1Pext$ b=-0.49 and $GMVSL_1S$ b=-0.56. However, the high bias can be addressed by using bias correction, a common practice in age-prediction which can benefit all models [51]. Although a comparison with other models would provide some important insights regarding the performance of SE, such a comparison would be unfair and potentially misleading as state-of-the-art models typically undergo an exhaustive model selection and feature engineering process, while here we did not perform any evaluation of different learning algorithms or parcellations schemes.

As expected, increasing the number of datasets used for training led to prediction performance improvements (figure 3). Especially for MAE, when using three datasets, was improved by 0.99 on average for GMV models, 0.42 for models that perform OOS predictions in level 0 and 0.45 for the other models (averaged across different train datasets and setup variations), compared to using one dataset for training. It is worth mentioning, however, that this was not the case for age bias, which demonstrates an unclear pattern in relation to the number of data sets. In fact, differences in bias increased with the number of datasets, likely due to differing age distributions across datasets. Since bias is influenced by the age distribution of the training subjects, its interpretation is challenging in a LOSO setup.

A coherent and robust association between chronological age and regional metrics can provide valuable biological insights into the healthy aging mechanism and consequently facilitate identification of brain regions susceptible to neurodegenerative and psychiatric diseases [26], [27]. To this end, we conducted a comparative analysis of the correlations between age and two metrics: the regional mean GMV and regional L0 age predictions. The results showed that the correlations were more pronounced for the L0 predictions, suggesting that the associated regional models provide robust biological insights and representation of healthy brain aging (Figures 4-5).

Additionally, this result underscores the ability of SE models to capture nuanced aging patterns in brain regions, offering a richer perspective than traditional analyses using mean GMV. Interestingly, subcortical areas showed a weak positive correlation with mean GMV, indicating that their volume increases with age which contradicts the current knowledge [5], [6], [7]. This effect could be, however, due to preprocessing artifacts [34]. Specifically, regions in thalamus had a positive GMV-real age correlation of r=0.28 and r=0.29, but showed high positive correlation of L0 predictions and real age, r=0.8 and r=0.62 respectively. Regional predictions from subcortical areas in SE showed a much higher correlation r=0.75, when the mean in subcortical areas for GMV was r=0.35 (mean of absolute values). This suggests that the L0 models were able to extract age-related information from individual voxels whereas their signal was diluted by the GMV averaging. Subcortical regions have been implicated in neurodegeneration processes related to Parkinson's disease and Alzheimer's disease [53], [54]. Therefore, appropriate modeling of GMV in those regions regarding healthy aging is essential for clinical application of brain age. In essence, SE effectively uses information from voxels that otherwise is "lost in the crowd" during averaging. Our correlations analysis between datasets suggests that the regional age predictions from SE models maintain a robust and stable association with age across all datasets, further underlining the reliability of the observed correlations (figures 5-6) . This result further supports the notion that the alignment of L0 regional age predictions with chronological age represents biological insights and is not contingent on specific characteristics of datasets. Such results can be further analyzed to provide improved clinical interpretations, though it is not in the scope of this work.

Identifying the origin dataset of individuals serves as an indirect measurement of privacy. A feature space that "hides" the dataset of origin also complicates subjects' identification. Dataset identification using L0 predictions of SE proved more challenging compared to using GMV ($ACC_{bal}$= 0.63 and $ACC_{bal}$= 0.87 respectively). Classification analyzes to predict the dataset suggest, to a certain extent, that the efficacy of SE concealing more information essential for identifying the dataset and consequently a subject's identity compared to GMV. This privacy preservation could be further improved by employing more complex models in the first level, such as tree-based models and deep neural networks. However, it is noteworthy that despite the high misclassification between subjects from IXI and eNKI, for CamCAN data and mostly for 1000Brains data the classification was accurate. A potential explanation for that can be the differences in data quality as well as the generally older population of these datasets. Nonetheless, this analysis can serve as a blueprint for other use cases helping the general radiomics community in developing privacy preserving and accurate methods. A potential improvement to the SE, aligned with privacy protection principles, would involve clinics sharing their first-level predictions and/or models with a remote central unit which could then train a comprehensive second-level model. This idea is similar to the approach used in federated learning in spirit and further architectures that combine it with SE could be developed [55].

Taken together, our comprehensive analyses allowed for a thorough understanding of the SE model generalizability and its ability to capture meaningful biological patterns. Future work could consider the application of more, and perhaps more suitable, models through a model selection process together with a more thorough hyperparameter tuning in both stacking ensemble levels. Such refinements will contribute to the overall effectiveness of the ensemble. At first, by improving the diversity and individual performance of base learners, and then by optimizing the combination of their predictions by the meta learner and refining the final prediction, leading to more robust and accurate models which will facilitate clinical application. As performance increased with training data size, including more sites in the training data we can further improve the performance of our SE model. Additionally, the final performance can be further enhanced by applying age-bias-correction in second level outcomes.

**Acknowledgements**

This study was partly supported by the Helmholtz-AI project BrainAge4AD (ZT-I-PF-5-163), and the Helmholtz Portfolio Theme "Supercomputing and Modelling for the Human Brain".

**Ethics statement**

Ethical approval and informed consent were obtained locally for each study covering both participation and subsequent data sharing. The ethics proposals for the use and retrospective analyses of the datasets were approved by the Ethics Committee of the Medical Faculty at the Heinrich-Heine-University Düsseldorf.

**Data/code availability statement**

The codes used for preprocessing, feature extraction and model training are available at https://github.com/juaml/stacking_ensemble_paper

**Conflict of interest**

KRP has a pending patent application pertaining to the SE methodology.

# Supplementary material

## Histograms of ages per dataset

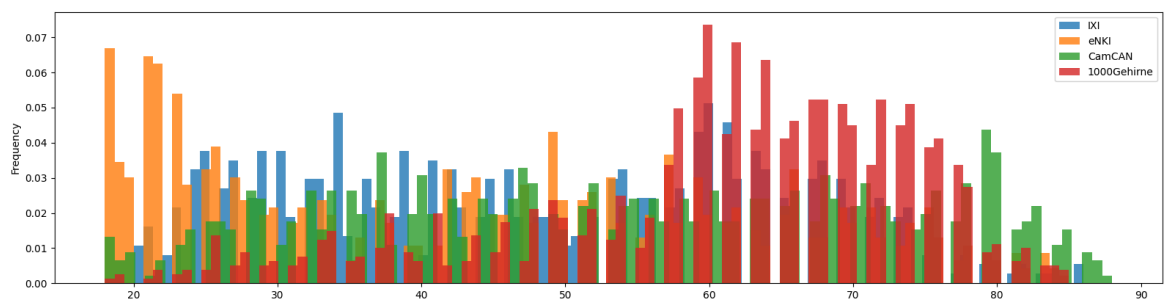

## Dataset classification results

| Model | Parameter | Values |
|---|---|---|
| Linear SVM | C | 0.01, 0.1, 1, 10 |
| SVM radial basis function kernel | C | 0.01, 0.1, 1, 10 |
| | γ | 1e-3, 1e-2, "scale", "auto" |
| Random forest | Maximum depth | 2, 3, 4 |
| Logistic regression | penalty | l1, l2 |

Sup. Table 1: Models and respective parameters that we tested for the task of dataset prediction given the two different feature spaces 1) regional age predictions from L0, and 2) regional mean GMV.

## Detailed results of setups

| Metric<br>Setup | MAE | r | $R^2$ | bias |
|---|---|---|---|---|
| $OOSPred_SL1_P$ | 4.75 | 0.93 | 0.86 | -0.44 |
| $OOSPred_SL1_S$ | 4.92 | 0.93 | 0.85 | -0.49 |
| $OOSPred_PL1_P$ | 4.95 | 0.93 | 0.85 | -0.52 |
| $PredL0_PL1_P$ | 5.06 | 0.94 | 0.84 | -0.43 |
| $PredL0_SL1_P$ | 5.12 | 0.93 | 0.84 | -0.49 |
| $PredL0_SL1_S$ | 5.15 | 0.93 | 0.84 | -0.49 |
| $GMV_PL1_P ext$ | 5.68 | 0.90 | 0.80 | -0.41 |
| $GMV_PL1_P$ | 6.22 | 0.89 | 0.76 | -0.45 |
| $GMV_SL1_S$ | 6.68 | 0.89 | 0.73 | -0.56 |

Supp. Table 2: Performance results for all setups in terms of MAE, Pearson's correlation, $R^2$ and bias. Models performing OOS in the test set in L0 had overall better performance, with the best one being the one using site predictions in L0 and pooling them to train L1 (MAE=4.75), however they demonstrated the highest bias.